\documentclass[11pt]{article}

\usepackage[final]{acl}

\usepackage{times}
\usepackage{latexsym}

\usepackage[T1]{fontenc}

\usepackage[utf8]{inputenc}

\usepackage{microtype}
\usepackage{amsmath}
\usepackage{multirow}

\usepackage{inconsolata}

\usepackage{graphicx}
\usepackage{subfigure}
\usepackage{paralist}
\usepackage{example}
\usepackage{booktabs}
\usepackage[ruled,vlined]{algorithm2e}
\usepackage[most]{tcolorbox}
\usepackage{amssymb}
\usepackage{makecell}
\usepackage{array}
\usepackage{hyperref}

\newtcolorbox{promptbox}[1]{
  colback=white,
  colframe=gray!70,
  coltitle=white,
  colbacktitle=gray!70,
  fonttitle=\bfseries,
  title=#1,
  arc=6pt,
  boxrule=1pt,
  left=6pt,
  right=6pt,
  top=6pt,
  bottom=6pt,
  breakable,        
}

\title{Debate to Align: Reliable Entity Alignment through Two-Stage Multi-Agent Debate}

\author{
 \textbf{Cunda Wang$^{1,2,3}$}\thanks{\ \ Equal contribution.},
 \textbf{Ziying Ma$^{1,2,3}$}\footnotemark[1],
 \textbf{Po Hu$^{1,2,3}$}\thanks{\ \ Corresponding authors.},
 \textbf{Weihua Wang$^{4, 5, 6}$}\footnotemark[2],
 \textbf{Feilong Bao$^{4, 5, 6}$}
\\
$^1$Hubei Provincial Key Laboratory of Artificial Intelligence and Smart Learning, \\Central China Normal University, Wuhan, China\\
$^2$School of Computer Science, Central China Normal University, Wuhan, China\\
$^3$National Language Resources Monitoring and Research Center for Network Media, \\Central China Normal University, Wuhan, China\\
$^4$College of Computer Science, Inner Mongolia University, Hohhot, China\\
$^5$National and Local Joint Engineering Research Center of Intelligent Information Processing \\ Technology for Mongolian, Inner Mongolia University, Hohhot, China\\
$^6$Inner Mongolia Key Laboratory of Multilingual Artificial Intelligence Technology, \\ Inner Mongolia University, Hohhot, China\\
\small \texttt{\{wangcunda, maziying, phu\}@mails.ccnu.edu.cn},
\small \texttt{\{wangwh, csfeilong\}@imu.edu.cn}
\\
}

\begin{document}
\maketitle
\begin{abstract}
Entity alignment (EA) aims to identify entities referring to the same real-world object across different knowledge graphs (KGs). Recent approaches based on large language models (LLMs) typically obtain entity embeddings through knowledge representation learning and use embedding similarity to identify an alignment-uncertain entity set. For each uncertain entity, a candidate entity set (CES) is then retrieved based on embedding similarity to support subsequent alignment reasoning and decision making. 
However, the reliability of the CES and the reasoning capability of LLMs critically affect the effectiveness of subsequent alignment decisions.
To address this issue, we propose AgentEA, a reliable EA framework based on multi-agent debate. AgentEA first improves embedding quality through entity representation preference optimization, and then introduces a two-stage multi-role debate mechanism consisting of lightweight debate verification and deep debate alignment to progressively enhance the reliability of alignment decisions while enabling more efficient debate-based reasoning. Extensive experiments on public benchmarks under cross-lingual, sparse, large-scale, and heterogeneous settings demonstrate the effectiveness of AgentEA~\footnote{Code is available at:  \url{https://github.com/eryueanran/AgentEA}.}.

\end{abstract}

\section{Introduction}

With the proliferation of heterogeneous knowledge graphs (KGs) from diverse sources and languages, entity alignment (EA)~\citep{openea} has become a fundamental task for identifying entities that refer to the same real-world object across different graphs. 
Well-constructed KGs can substantially benefit various downstream tasks~\citep{kbqaldy1,tkge, edurh, rhtkga,rhqa}.
Early EA approaches~\citep{mtranse,wang2024unifying} based on knowledge representation learning (KRL) typically employ translation-based models or graph neural networks to project entities into a shared embedding space and predict correspondences via embedding similarity. However, KRL-based alignment only captures proximity in the embedding space and fails to explicitly model semantic, structural, and attribute-level consistency, while lacking the reasoning capability required to handle complex ambiguity. As a result, KRL-based methods often struggle to reliably distinguish semantically similar but non-aligned entities, limiting their effectiveness in real-world, complex EA scenarios.

\begin{figure}
    \centering
    \includegraphics[width=1\linewidth]{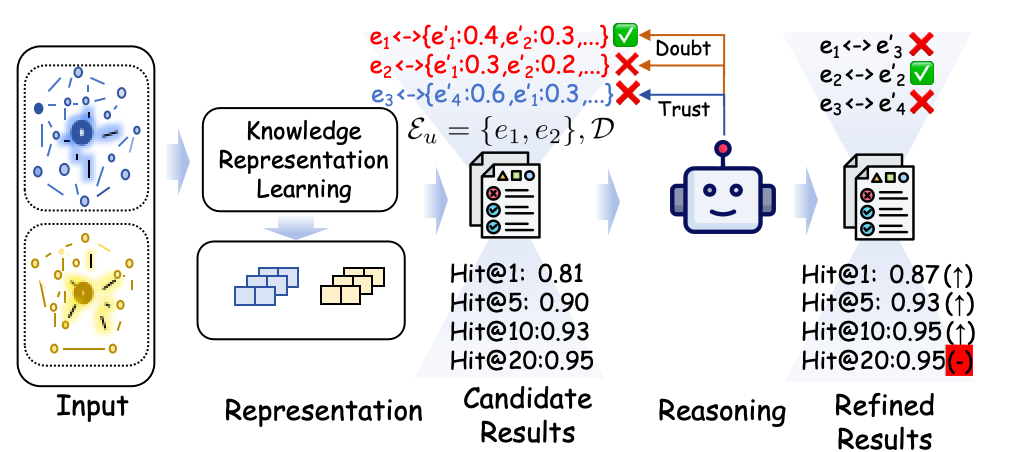}
    \caption{Overview of an LLM-based EA framework. Entities with a small Top-1/Top-2 similarity gap (e.g., $e_1$ and $e_2$) are forwarded to LLM-based reasoning, whereas those with a large gap are directly aligned (e.g., $e_3$).}
    \label{fig:head}
\end{figure}

Recently EA methods based on large language models (LLMs) have attracted increasing attention~\citep{llmsurvey}. Benefiting from their strong capabilities in semantic understanding, contextual modeling, and complex reasoning, existing studies leverage LLMs to model entity names, descriptions, and attribute information, thereby compensating for the limited semantic expressiveness of traditional KRL methods. 
As illustrated in Figure~\ref{fig:head}, a typical LLM-based EA framework first learns entity embeddings via KRL and selects an alignment-uncertain entity set (AES) $\mathcal{E}_u$ based on embedding similarity. For each $e_i \in \mathcal{E}_u$, a candidate entity set (CES) $\mathcal{D}(e_i)$ is then retrieved from the target KG, 
over which the LLM conducts single-pass, candidate-conditioned reasoning to produce alignment decisions.

However, the upper bound of LLM-based EA is jointly constrained by the quality of candidate retrieval and the inherent limitations of single-LLM reasoning. As shown in Figure~\ref{fig:head}, constructing the uncertain entity set $\mathcal{E}_u$ based on heuristic thresholds over the Top-1/Top-2 embedding similarity gap is unreliable, as unstable similarity distributions can mischaracterize alignment ambiguity, causing genuinely ambiguous entities (e.g., $e_3$) to be excluded from reasoning while incorrect alignments remain unchallenged.
More critically, alignment decisions in existing LLM-based EA methods are typically derived from a single, unchallenged reasoning trajectory. Without explicit support for alternative hypotheses or counter-evidence~\cite{mad}, the model cannot revise its conclusions under incomplete or misleading evidence, leading to systematic failures such as rejecting a true alignment (e.g., $e_1$).
Overall, LLM-based alignment decisions for each entity $e_i$ are strictly conditioned on its candidate set $\mathcal{D}(e_i)$. If the true aligned entity is absent, the correct alignment becomes unrecoverable regardless of the model’s reasoning capability. Consequently, the Hit@1 performance of LLM-based EA is inherently upper-bounded by the Hit@20 performance of KRL-based candidate retrieval.

To address the aforementioned issues, we propose AgentEA, a reliable entity alignment framework based on multi-agent debate. AgentEA improves EA performance by jointly optimizing entity embeddings and incorporating structured multi-agent reasoning. Specifically, we introduce an entity representation preference optimization module that enhances embedding quality by fine-tuning a large language model with Direct Preference Optimization (DPO). 
The training corpus is constructed using name similarity-based negative samples, degree-aware neighborhood negative samples, and seed entity pairs as positive samples.
Furthermore, we design a two-stage multi-role debate mechanism consisting of lightweight debate verification and deep debate alignment, which jointly improves both the effectiveness and efficiency of multi-agent EA. 
The lightweight debate verification stage employs three agents to reason over entities in $\mathcal{E}_u$ and perform consistency checking. Entity pairs that exhibit disagreement, as well as low-confidence entities, are then forwarded to the deep debate alignment stage, where six agents debate from multiple perspectives to enhance the reliability of alignment decisions.

In general, our main contributions are as follows:
\begin{compactitem}
\item We propose AgentEA, which, to the best of our knowledge, is the first framework to improve EA performance through multi-agent debate.
\item We design two strategies for constructing hard negative samples and leverage Direct Preference Optimization to fine-tune a LLM, significantly enhancing entity embedding quality.
\item We design a two-stage multi-role debate framework that improves both the reliability and efficiency of alignment decisions.
\item Extensive experiments on multiple datasets demonstrate the effectiveness of AgentEA.

\end{compactitem}

\section{Related Work}
In this section, we categorize existing EA methods into two main groups: KRL-based alignment methods and LLM–based alignment methods. In addition, we review related work on multi-agent debate across different domains.

\textbf{KRL-based Alignment Methods.}
This category of methods primarily encodes entities using translation-based models or graph neural networks (GNNs) and map entities from different knowledge graphs into a shared vector space. For example, MTransE~\cite{mtranse}, IPTransE~\cite{iptranse}, and TransEdge~\cite{transedge} model triples under the translation assumption, while subsequent approaches introduce graph neural networks with gating mechanisms (e.g., HGCN~\cite{hgcn}, AliNet~\cite{alinet}, RHGN~\cite{rhgn}, and LGEA~\cite{lgea}) to better capture structural information and alleviate representation over-smoothing. However, these KRL-based methods primarily rely on structure-driven representation learning and remain limited in scenarios involving semantic ambiguity or incomplete graph structures.

\begin{figure*}
    \centering
    \includegraphics[width=1\linewidth]{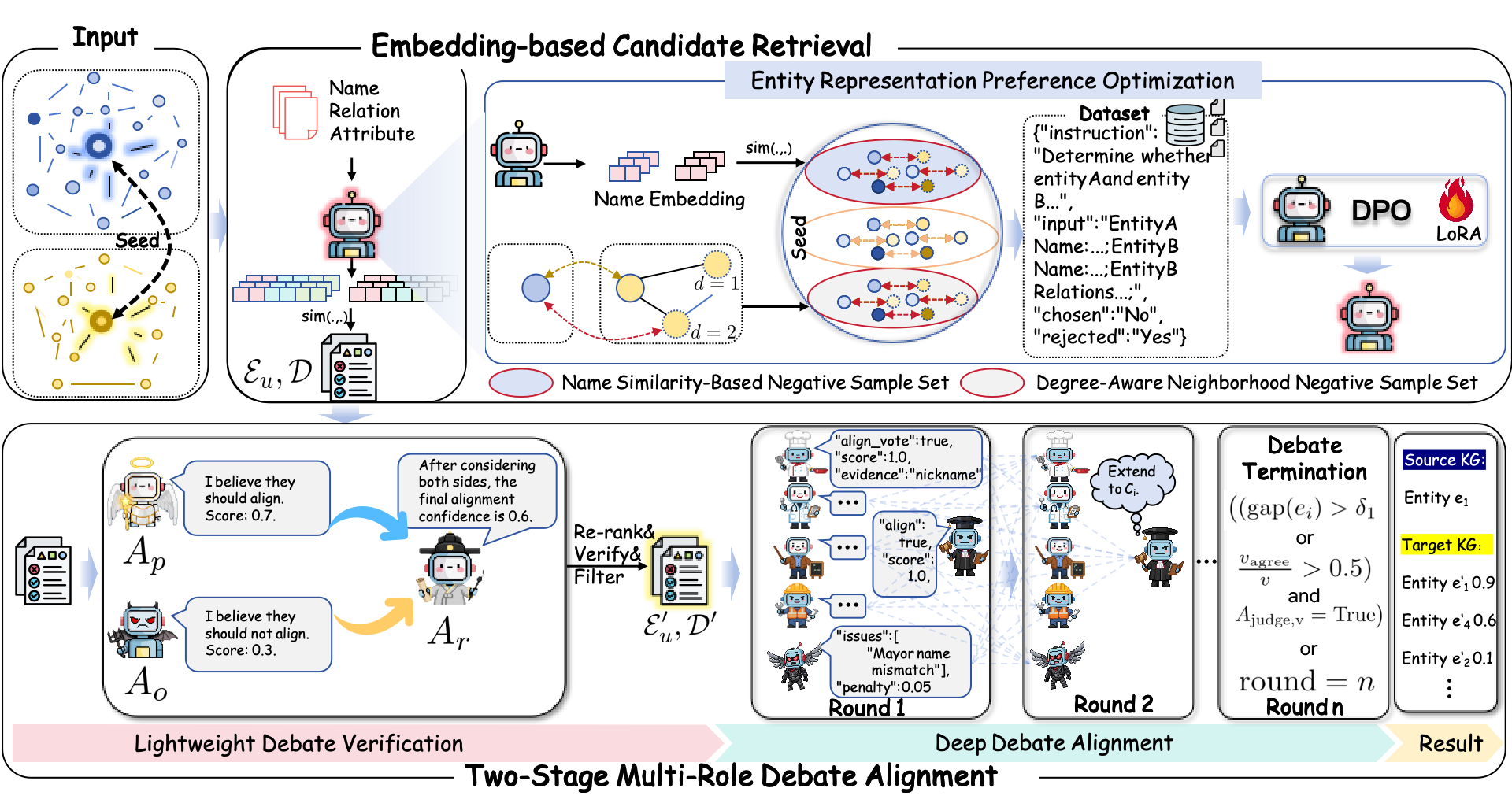}
    \caption{The overview framework of our proposed AgentEA, which consists of two main components: (1) embedding-based candidate retrieval and (2) two-stage multi-role debate alignment.}
    \label{fig:model}
\end{figure*}

\textbf{LLM-based Alignment Methods.}
Recently, these methods typically take candidate entities generated by KRL-based approaches as input and leverage the semantic understanding and reasoning capabilities of LLMs to re-evaluate entity pairs. To enhance LLMs’ awareness of knowledge graph structure, ChatEA~\cite{chatea} and MM-ChatAlign~\cite{MM-ChatAlign} transform entity neighborhoods and relations into code-like representations as model inputs. EasyEA~\cite{easyea} and ProLEA~\cite{prolea} activate LLMs’ rich background knowledge through knowledge summarization, while Seg-Align~\cite{seg-align} and LLM4EA~\cite{llm4ea} explore the zero-shot capabilities of LLMs for EA. 
However, most of these methods rely on a single LLM to directly make alignment decisions and lack explicit multi-perspective verification mechanisms, making them vulnerable to uncertainty and bias in cases of high semantic similarity or conflicting evidence, and thus limiting the stability and reliability of alignment results.

\textbf{Multi-Agent Debate.}
Multi-Agent Debate (MAD) improves reasoning quality by integrating diverse perspectives from multiple agents and has been shown to effectively mitigate hallucination in LLMs~\cite{MAD1}. Existing studies on organizing multi-agent collaborative reasoning mainly explore two directions: role design and debate structure.
On the one hand, role heterogeneity is introduced to enhance viewpoint diversity, as exemplified by methods such as MRBalance~\cite{MRBalance}, MAD-AWSD~\cite{MADAWSD}, DRAG~\cite{DRAG}, and M-MAD~\cite{mmad}.
On the other hand, various debate structures are proposed to improve the stability and self-correction ability of multi-round reasoning, including BELLE’s~\cite{BELLE} fast–slow debater mechanism and the self-reflective reasoning framework proposed by~\citet{Ki}.
Compared with existing approaches, AgentEA introduces a two-stage multi-role debate framework that progresses from lightweight debate verification to deep debate alignment. Through lightweight debate verification of the alignment-uncertain entity set, AgentEA significantly improves the efficiency and performance of deep debate alignment while maintaining reliable alignment decisions.

\section{Method}
As illustrated in Figure~\ref{fig:model}, we first introduce the entity representation preference optimization module for embedding-based candidate retrieval, followed by the two-stage multi-role debate alignment framework, which consists of lightweight debate verification and deep debate alignment.

\subsection{Embedding-based Candidate Retrieval}
Following EasyEA~\cite{easyea}, we first summarize relational triples and attribute triples separately and employ a LLaMA-based encoder to obtain representations for entity names, relations, and attributes. These representations are then concatenated and fused to produce the final embedding for each entity.

\subsubsection{Set Initialization}
Based on the learned entity embeddings, for each source entity $e_i \in \mathcal{E}_s$, we compute similarity scores with entities in the target KG to construct its target entity candidate set, which is formally defined as
\begin{equation}
\mathcal{D}(e_i) = \left\{ (e'_j, s_{i,j}) \mid e'_j \in \mathcal{E}_t \right\},
\end{equation}
where $s_{i,j}$ denotes the similarity score between the source entity $e_i$ and the target entity $e'_j$.

\textbf{Initialization of the Alignment-Uncertain Entity Set.}
A source entity is regarded as uncertain if the similarity gap between its top two candidate entities is smaller than a predefined threshold:
\begin{equation}
\mathcal{E}_u = \left\{ e_i \;\middle|\; s_{i,1} - s_{i,2} < \delta_1 \right\},
\end{equation}
where $s_{i,1}$ and $s_{i,2}$ denote the highest and second-highest similarity scores for the source entity $e_i$, respectively, and $\delta_1$ is a predefined threshold.

\textbf{Initialization of the Candidate Entity Set.}
For each uncertain source entity $e_i \in \mathcal{E}_u$, the corresponding target entity candidate set $\mathcal{D}(e_i)$ is then used as input for subsequent multi-agent debate alignment.

\subsubsection{Entity Representation Preference Optimization}
For each source entity $e_i$, whether the true aligned entity is included in its candidate set $\mathcal{D}(e_i)$ is a necessary condition for making correct decisions in subsequent reasoning stages. To improve the quality of entity representations in the candidate retrieval stage, we propose an {Entity Representation Preference Optimization} method, which fine-tunes a LLaMA-based encoder via {Direct Preference Optimization} (DPO)~\cite{DPO} to obtain more discriminative entity embeddings.

The training data are constructed from the three cross-lingual DBP15K datasets and are formally defined as
\begin{equation}
\mathcal{T} = \mathcal{S} \cup \mathcal{N}_{\text{name}} \cup \mathcal{N}_{\text{nbr}},
\end{equation}
where $\mathcal{S}$ denotes the set of seed entity pairs, $\mathcal{N}_{\text{name}}$ represents the set of name similarity--aware negative entity pairs, and $\mathcal{N}_{\text{nbr}}$ corresponds to the set of degree-aware neighborhood negative entity pairs.

\textbf{Name Similarity--based Negative Samples.}
In cross-lingual entity alignment (EA), entities with similar names often exhibit high semantic overlap, making them particularly prone to misalignment. To address this issue, we propose a {name similarity--aware negative sampling strategy} that constructs hard negative samples based on misleading name similarity.
Specifically, we first translate all entity names in the DBP15K training set into English and encode them using a pretrained language model to obtain name representations. We then compute pairwise name similarities between entities. For each source entity $e_i$, we select the entity that is most similar in name but not its ground-truth counterpart as the negative sample:
\begin{equation}
e_i^{N-} = \arg\max_{e_j \neq e_i^+}
\cos \big( \mathbf{h}_i^{\text{name}}, \mathbf{h}_j^{\text{name}} \big),
\end{equation}
where $e_i^+$ denotes the ground-truth aligned entity of $e_i$.
The resulting name similarity--aware negative sample set is denoted as $\mathcal{N}_{\text{name}} = \{(e_i, e_i^{N-})\}$.


\textbf{Degree-aware Neighborhood Negative Samples.}
Entities with higher degrees in their local neighborhoods often exhibit more complex structural contexts, which are more likely to introduce structural noise during entity alignment. To address this issue, we propose a {degree-aware neighborhood negative sampling strategy} that constructs hard negative samples based on neighborhood degree information.
Specifically, for a source entity $e_i$, we first define its neighborhood entity set $\mathrm{Nbr}(e_i)$ and compute the degree $\deg(e_k)$ for each neighboring entity $e_k \in \mathrm{Nbr}(e_i)$. We then select the neighboring entity with the highest degree as the negative sample:
\begin{equation}
e_i^{D-} = \arg\max_{e_k \in \mathrm{Nbr}(e_i)} \deg(e_k).
\end{equation}
The resulting degree-aware neighborhood negative sample set is denoted as $\mathcal{N}_{\text{nbr}} = \{(e_i, e_i^{D-})\}$.


\subsection{Two-Stage Multi-Role Debate Alignment}
The two-stage multi-role debate alignment consists of {Lightweight Debate Verification} (LDV) and {Deep Debate Alignment} (DDA), where a progressive debate structure balances efficiency and reliability, and role heterogeneity enables multi-perspective reasoning to enhance the robustness of entity alignment decisions.

\subsubsection{Lightweight Debate Verification}
LDV aims to efficiently verify and filter candidate entities through low-cost multi-agent reasoning. Specifically, this stage employs a single-round debate mechanism composed of three agents to re-rank the candidate entity set $\mathcal{D}(e_i)$ for a source entity $e_i$, and performs consistency verification and filtering based on the debate outcomes, thereby updating the {Uncertain Source Entity Set} $\mathcal{E}_u$. The updated $\mathcal{E}'_u$ retains only those source entities whose alignment decisions remain disputed and are more difficult to distinguish. This design focuses computational resources on challenging cases in the subsequent DDA stage, improving the reliability of final alignment decisions while maintaining overall efficiency.

To reduce the reasoning cost of large language models, the LDV stage first compresses entity information. Specifically, we adopt a frequency-based strategy to extract key attributes and relations while filtering out low-value information. After compression, the entity description typically retains less than $15\%$ of the original input tokens, while still preserving the core semantic cues required to distinguish candidate entities.

On this basis, the LDV stage adopts a lightweight three-agent architecture with a single-round debate, consisting of an {alignment proponent} $A_p$, an {alignment opponent} $A_o$, and an {alignment referee} $A_r$. The agent $A_p$ provides supporting evidence in favor of entity alignment, $A_o$ presents conflicting evidence against alignment, and $A_r$ integrates both perspectives to make a final judgment and re-rank the candidate entities accordingly.

Overall, the LDV stage can be formally expressed as:
\begin{equation}
\big(\mathcal{D}'(e_i),\, \mathcal{E}_u'\big)
=
f_{\mathrm{LDV}}\!\left(e_i,\, \mathcal{D}(e_i)\right), e_i\in\mathcal{E}_u
\end{equation}
where taking the source entity $e_i$ and its initial candidate set $\mathcal{D}(e_i)$ as input, and producing the re-ranked candidate set $\mathcal{D}'(e_i)$ together with the updated uncertain source entity set $\mathcal{E}_u'$.

\subsubsection{Deep Debate Alignment}
The DDA stage introduces a multi-role, multi-round debate structure to perform fine-grained differentiation over the candidate entity set $\mathcal{D}'(e_i)$ for a source entity $e_i\in\mathcal{E}_u'$. 

\textbf{Debate Structure.}
In the first debate round, all agents except the judge independently analyze the candidate entities based on the original entity information. The judge agent $A_{judge}$ aggregates the scores from all agents and determines whether the early stopping condition is satisfied.  
In the subsequent rounds ($2$ to $n\!-\!1$), an interactive error-correction strategy is adopted. All agents are allowed to access the complete debate outputs from the previous round. The attacker agent $A_{attack}$ integrates the scores from specialized agents and applies more fine-grained penalties to high-risk candidates, while $A_{judge}$ aggregates updated scores and decides whether to trigger candidate set expansion.  
In the final round, a forced termination mechanism is applied, where $A_{judge}$ produces the final alignment score based on the accumulated debate results from all rounds.

\textbf{Agent Roles.}
DDA stage consists of four types of specialized agents that analyze candidate entities from different perspectives, including entity names, types, attributes, and neighborhood structures. Each specialized agent $A_i \in \mathcal{A}_s$ is equipped with a corresponding prompt template $P_{A_i}$.  
$A_{attack}$ acts as a reflective role that challenges the judgments of other agents by identifying conflicting evidence and potential risks. $A_{judge}$ is responsible for aggregating evidence and scores across multiple debate rounds, deciding whether to expand the candidate set or terminate the debate, and finally producing the alignment decision.
Details are given in Appendix~\ref{role}.

\textbf{Progressive Candidate Set Expansion.}
Since conducting multi-role debates over all candidates in $\mathcal{D}'(e_i)$ incurs substantial computational cost, we adopt a progressive candidate set expansion strategy. Specifically, multiple candidate subsets are constructed by ranking candidates according to scores:
$
\mathcal{D}_k(e_i)
=
\{(e'_j, s_{i,j})|j \le k\},
 k \in \{5,10,15,20\}.
$
Candidate set expansion is triggered when the following conditions are satisfied:
$
\text{s$_{i,1}$}(e_i) < \delta_2 
\;\land\;
\frac{v_{\text{agree}}}{v} \le  0.5
\;\land\;
A_{\text{judge,v}} = \text{False},
$
where
$
\text{s$_{i,1}$}(e_i)
$
denotes that the similarity score of the Top-1 candidate entity corresponding to $e_i$, $\tfrac{v_{\text{agree}}}{v}$ represents the proportion of votes supporting alignment, and $A_{\text{judge,v}}\in\{\text{True},\text{False}\}$ indicates the alignment judgment made by the judge agent.

\textbf{Termination Conditions.}
If a reliable alignment decision is reached in early rounds, the debate terminates early; otherwise, it is forcibly terminated upon reaching the maximum number of rounds $n$. The termination condition is defined as:
$
\big((\text{gap}(e_i) > \delta_1 
\;\lor\;
\tfrac{v_{\text{agree}}}{v} > 0.5)
\;\land\;
A_{\text{judge,v}} = \text{True}\big)
\;\lor\;
\text{round} = n,
$
where
$
\text{gap}(e_i)=s_{i,1}-s_{i,2},
$
denotes the similarity gap between the top two candidates in the current candidate set.

\section{Experiments}

In this section, we conduct a systematic evaluation of {AgentEA} on multiple entity alignment benchmark datasets. Our experiments are designed to answer the following research questions:
\begin{compactitem}
\item \textbf{RQ1:} Can AgentEA effectively overcome the limitations of existing  methods?
\item \textbf{RQ2:} Does AgentEA significantly improve the representation and discriminative capability of entity embeddings?
\item \textbf{RQ3:} How effective are the key components of AgentEA?
\item \textbf{RQ4:} How does AgentEA perform in terms of computational efficiency and reasoning cost?
\item \textbf{RQ5:} How does the choice of different LLMs affect the performance of AgentEA?
\item \textbf{RQ6:} How do key hyper-parameter settings influence alignment performance?
\item \textbf{RQ7:} Why does MAD outperform single-LLM reasoning in EA? (See~\ref{case})
\end{compactitem}

\begin{table*}[t]
\centering
\setlength{\tabcolsep}{4pt} 
\begin{tabular}{l|ccc|ccc|ccc}
\hline
\multirow{2}{*}{\textbf{Models}} &
\multicolumn{3}{c|}{\textbf{DBP15K$_{\text{ZH-EN}}$}} &
\multicolumn{3}{c|}{\textbf{DBP15K$_{\text{JA-EN}}$}} &
\multicolumn{3}{c}{\textbf{DBP15K$_{\text{FR-EN}}$}} \\
& Hits@1 & Hits@10 & MRR & Hits@1 & Hits@10 & MRR & Hits@1 & Hits@10 & MRR \\
\hline
\multicolumn{10}{c}{\textit{Knowledge Representation Learning-based Entity Alignment Methods}} \\
\hline
MTransE    & 0.308 & 0.614 & 0.364 & 0.279 & 0.575 & 0.349 & 0.247 & 0.577 & 0.360 \\
GCN-Align  & 0.413 & 0.744 & 0.549 & 0.399 & 0.745 & 0.546 & 0.411 & 0.772 & 0.530 \\
BootEA     & 0.629 & 0.848 & 0.703 & 0.622 & 0.854 & 0.701 & 0.653 & 0.874 & 0.731 \\
RDGCN      & 0.708 & 0.846 & 0.746 & 0.767 & 0.895 & 0.812 & 0.873 & 0.950 & 0.901 \\
Dual-AMN   & 0.861 & 0.964 & 0.901 & 0.892 & 0.978 & 0.925 & 0.954 & 0.994 & 0.970 \\
RHGN   & 0.379 & 0.641 & 0.471 & 0.389 & 0.649 & 0.482 & 0.406 & 0.695 & 0.507 \\
SCMEA   & 0.908 & 0.979 & 0.935 & 0.934 & 0.987 & 0.954 & 0.982 & 0.998 & 0.988 \\
UniEA   & 0.486 & 0.788 & 0.590 & 0.473 & 0.801 & 0.585 & 0.504 & 0.836 & 0.618 \\
\hline
\multicolumn{10}{c}{\textit{Large Language Models-based Entity Alignment Methods}} \\
\hline
LLMEA      & 0.898 & 0.923 & --    & 0.911 & 0.946 & --    & 0.957 & 0.977 & --    \\
Seg-Align  & 0.953 & --    & --    & 0.907 & --    & --    & 0.987 & --    & --    \\
ChatEA     & .951    & .965    & .970    & .972    & .991    & .980    & 0.990 & 1.000 & 0.995 \\

{EasyEA}*  & {0.863} & {0.977} & {-} &
                  {0.912} & {0.991} & {-} &
                  {0.927} & {0.998} & {-} \\
{ProLEA}  & {-} & {-} & {-} &
                  {-} & {-} & {-} &
                  {0.993} & {1.000} & {0.997} \\
{AdaCoAgentEA}  & {-} & {-} & {-} &
                  {-} & {-} & {-} &
                  {0.964} & {0.982} & {0.971} \\

\hline
\textbf{AgentEA} & \textbf{0.970} & \textbf{0.981} & \textbf{0.974} &
                  \textbf{0.984} & \textbf{0.992} & \textbf{0.987} &
                  \textbf{0.996} & \textbf{1.000} & \textbf{0.998} \\
\hline
\end{tabular}
\caption{Main experiment results on DBP15K dataset. * denotes results reproduced using publicly available code.}
\label{tab:dbp15k-results}
\end{table*}

\begin{table*}[t]
\centering
\setlength{\tabcolsep}{2.5pt} 
\begin{tabular}{l|ccc|ccc|ccc|ccc}
\hline
\multirow{2}{*}{\textbf{Models}} &
\multicolumn{3}{c|}{\textbf{ICEWS-WIKI}} &
\multicolumn{3}{c|}{\textbf{ICEWS-YAGO}} &
\multicolumn{3}{c|}{\textbf{DWY$_{\text{DBP-WIKI}}$}} &
\multicolumn{3}{c}{\textbf{DWY$_{\text{DBP-YAGO}}$}} \\
& H@1 & H@10 & MRR & H@1 & H@10 & MRR & H@1 & H@10 & MRR & H@1 & H@10 & MRR \\
\hline
\multicolumn{13}{c}{\textit{Knowledge Representation Learning-based Entity Alignment Methods}} \\
\hline
MTransE   & 0.021 & 0.158 & 0.068 & 0.012 & 0.084 & 0.040 & 0.281 & 0.520 & 0.363 & 0.252 & 0.493 & 0.334 \\
GCN-Align & 0.046 & 0.184 & 0.093 & 0.017 & 0.085 & 0.038 & 0.506 & 0.772 & 0.600 & 0.597 & 0.838 & 0.682 \\
BootEA    & 0.072 & 0.275 & 0.139 & 0.020 & 0.120 & 0.056 & 0.748 & 0.898 & 0.801 & 0.761 & 0.894 & 0.808 \\
RDGCN     & 0.064 & 0.202 & 0.096 & 0.029 & 0.097 & 0.042 & 0.623 & 0.805 & 0.684 & 0.936 & 0.973 & 0.950 \\

Dual-AMN  & 0.083 & 0.281 & 0.145 & 0.031 & 0.144 & 0.068 & 0.869 & 0.969 & 0.908 & 0.907 & 0.981 & 0.935 \\
\hline
\multicolumn{13}{c}{\textit{Large Language Models-based Entity Alignment Methods}} \\
\hline
LLM4EA  & - & - & - & - & - & - & 0.898 & 0.979 & 0.929 & 0.979 & 0.996 & 0.985 \\
ChatEA  & 0.880 & 0.945 & 0.912 & 0.935 & 0.955 & 0.944 & 0.995 & 1.000 & 0.998 & - & - & - \\

{EasyEA}*  & {0.864} & {0.938} & {-} &
                  {0.906} & {0.961} & {-} &
                  {0.952} & {0.987} & {-} &
                  {0.954} & {0.981} & {-} \\
{ProLEA}  & {0.964} & {0.975} & {0.972} &
                  {0.969} & {0.981} & {0.977} &
                  {0.997} & {1.000} & {0.998} &
                  {-} & {-} & {-} \\
{AdaCoAgentEA}  & {0.884} & {0.948} & {0.905} &
                  {0.862} & {0.930} & {0.891} &
                  {0.957} & {0.984} & {0.975} &
                  {-} & {-} & {-} \\
\hline
\textbf{AgentEA} & \textbf{0.973} & \textbf{0.986} & \textbf{0.984} &
                  \textbf{0.982} & \textbf{0.991} & \textbf{0.989} &
                  \textbf{1.000} & \textbf{1.000} & \textbf{1.000} &
                  \textbf{0.987} & \textbf{0.998} & \textbf{0.992} \\
\hline
\end{tabular}
\caption{Results on ICEWS-WIKI, ICEWS-YAGO, DWY$_{\text{DBP-WIKI}}$ and DWY$_{\text{DBP-YAGO}}$  datasets.}
\label{tab:dual-results}
\end{table*}

\subsection{Experimental Settings}

We briefly introduce the datasets, evaluation metrics, baselines, and implementation details used in our experiments. See Appendix~\ref{f}, \ref{l}, \ref{d}, and \ref{m} for experimental details.

\textbf{Datasets.}
We conduct experiments on 11 widely used entity alignment benchmark datasets, including DBP15K (ZH--EN, JA--EN, FR--EN)~\cite{bert-int}, ICEWS (WIKI, YAGO)~\cite{ICEWS}, DWY (DBP--WIKI, DBP--YAGO)~\cite{DWY}, and SRPRS (EN--DE, EN--FR, DBP--YAGO, DBP--WIKI)~\cite{SRPRS}.

\textbf{Evaluation Metrics.}
We adopt Hit@$K$ and Mean Reciprocal Rank (MRR) as evaluation metrics. Higher values indicate better performance.

\textbf{Implementation Details.}
All experiments are conducted on a single NVIDIA A100 GPU.  
In the embedding-based candidate retrieval stage, we fine-tune the open-source LLM \texttt{LLaMA3-8B-Instruct} using Direct Preference Optimization (DPO) implemented via the \texttt{llama-factory}~\citep{llamafactory}. The similarity threshold is set to $\delta_1 = 0.05$.  
In the two-stage multi-role debate alignment module, we adopt \texttt{gpt-3.5-turbo} as the backbone large language model. The maximum number of debate rounds is set to 3, with $\delta_2 = 0.5$.

\textbf{Baselines.}
We categorize existing EA methods into KRL--based and LLM--based methods.  
\textbf{KRL-based methods} include MTransE~\citeyear{mtranse}, GCN-Align~\citeyear{gcn-align}, BootEA~\citeyear{bootea}, RDGCN~\citeyear{rdgcn}, Dual-AMN~\citeyear{dual-amn}, MuGNN~\citeyear{mugnn}, NAEA~\citeyear{naea}, KECG~\citeyear{kecg}, RSN4EA~\citeyear{rsn4ea}, TransEdge~\citeyear{transedge}, and MRAEA~\citeyear{mraea}, RHGN~\citeyear{rhgn}, SCMEA~\citeyear{scmea}, UniEA~\citeyear{wang2024unifying}.  
\textbf{LLM-based methods} include LLM4EA~\citeyear{llm4ea}, LLMEA~\citeyear{LLMEA}, ChatEA~\citeyear{chatea}, Seg-Align~\citeyear{seg-align}, EasyEA~\citeyear{easyea}, AdaCoAgentEA~\citeyear{AdaCoAgentEA} and ProLEA~\citeyear{prolea}.

\subsection{Main Results (RQ1)}
Tables~\ref{tab:dbp15k-results},~\ref{tab:dual-results}, and~\ref{tab:srprs-results} report the overall performance of AgentEA on multiple EA benchmark datasets. The results show that AgentEA consistently outperforms existing EA methods across all evaluation metrics, demonstrating its effectiveness in complex alignment scenarios and indicating its ability to overcome the performance bottlenecks of current LLM-based alignment approaches. Notably, on the DWY (DBP–WIKI) and SRPRS (DBP–YAGO) datasets, AgentEA achieves perfect scores of 1.00 on multiple metrics, highlighting its strong alignment stability and reliability.

Overall, LLM-based entity alignment methods generally outperform traditional KRL-based approaches, highlighting the advantages of large language models in semantic understanding and complex reasoning. However, most existing LLM-based methods rely on a single model for direct alignment decisions and lack explicit verification mechanisms, which can lead to instability under semantic ambiguity or conflicting evidence. In contrast, AgentEA combines high-quality entity embeddings with a two-stage multi-role debate framework to mitigate single-model reasoning instability and produce more reliable alignment decisions.


\begin{table}[t]
\centering
\setlength{\tabcolsep}{2.3pt} 
\begin{tabular}{l|cc|cc}
\hline
\multirow{2}{*}{\textbf{Settings}} &
\multicolumn{2}{c|}{\textbf{DBP$_{\text{FR-EN}}$}} & \multicolumn{2}{c}{\textbf{SRPRS$_{\text{EN-DE}}$}} \\
& Hits@1 & MRR & Hits@1 & MRR \\
\hline
AgentEA              & 0.996 & 0.998 & 0.977 & 0.980 \\
- only DPO           & 0.976 & 0.986 & 0.948 & 0.963 \\
- w/o DPO            & 0.859 & 0.889 & 0.862 & 0.888 \\
- w/o DN         & 0.938 & 0.955 & 0.928 & 0.945 \\
- w/o NN      & 0.960 & 0.976 & 0.945 & 0.960 \\
- w DPO+random       & 0.914 & 0.943 & 0.904 & 0.928 \\
\hline
\end{tabular}
\caption{Ablation experiments on DBP15K$_{\text{FR-EN}}$ and SRPRS$_{\text{EN-DE}}$ datasets during the candidate entity generation phase.}
\label{tab:ablation1}
\end{table}

\begin{figure}[ht]
\centering 

\subfigure[w/o DPO]{
\label{Fig:oDPO}
\includegraphics[width=.75\linewidth]{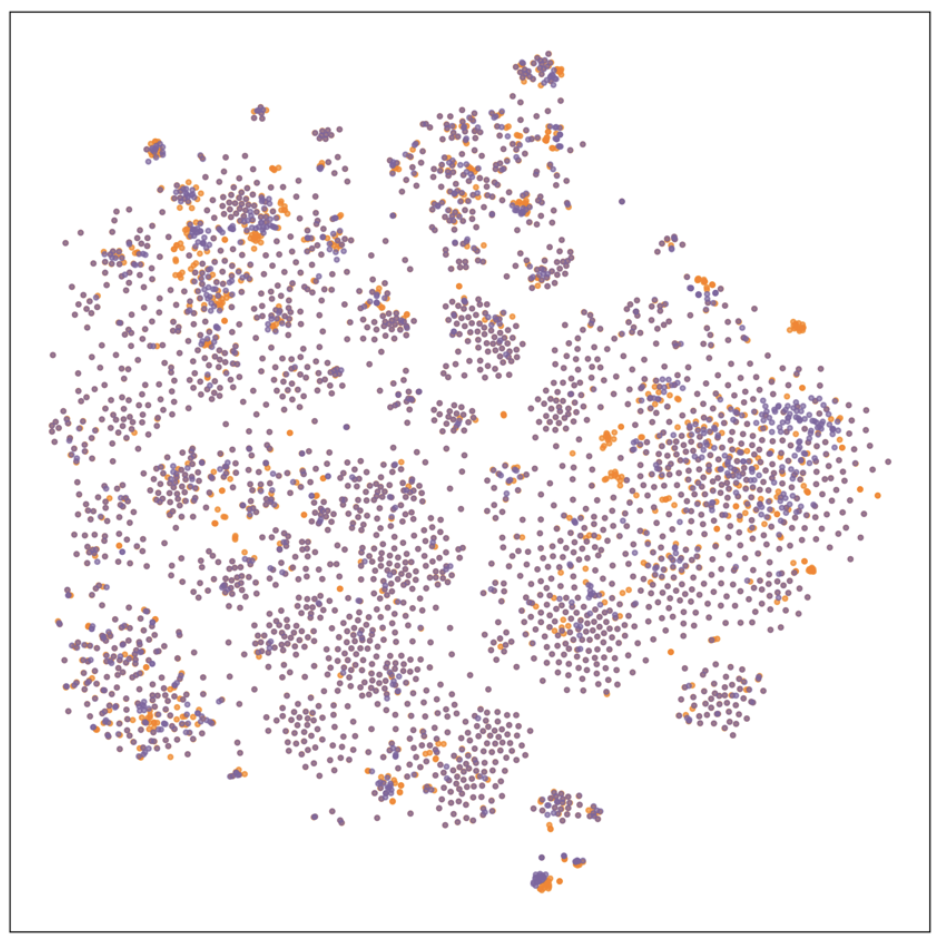}}
\subfigure[only DPO]{
\label{Fig:DPO}
\includegraphics[width=.75\linewidth]{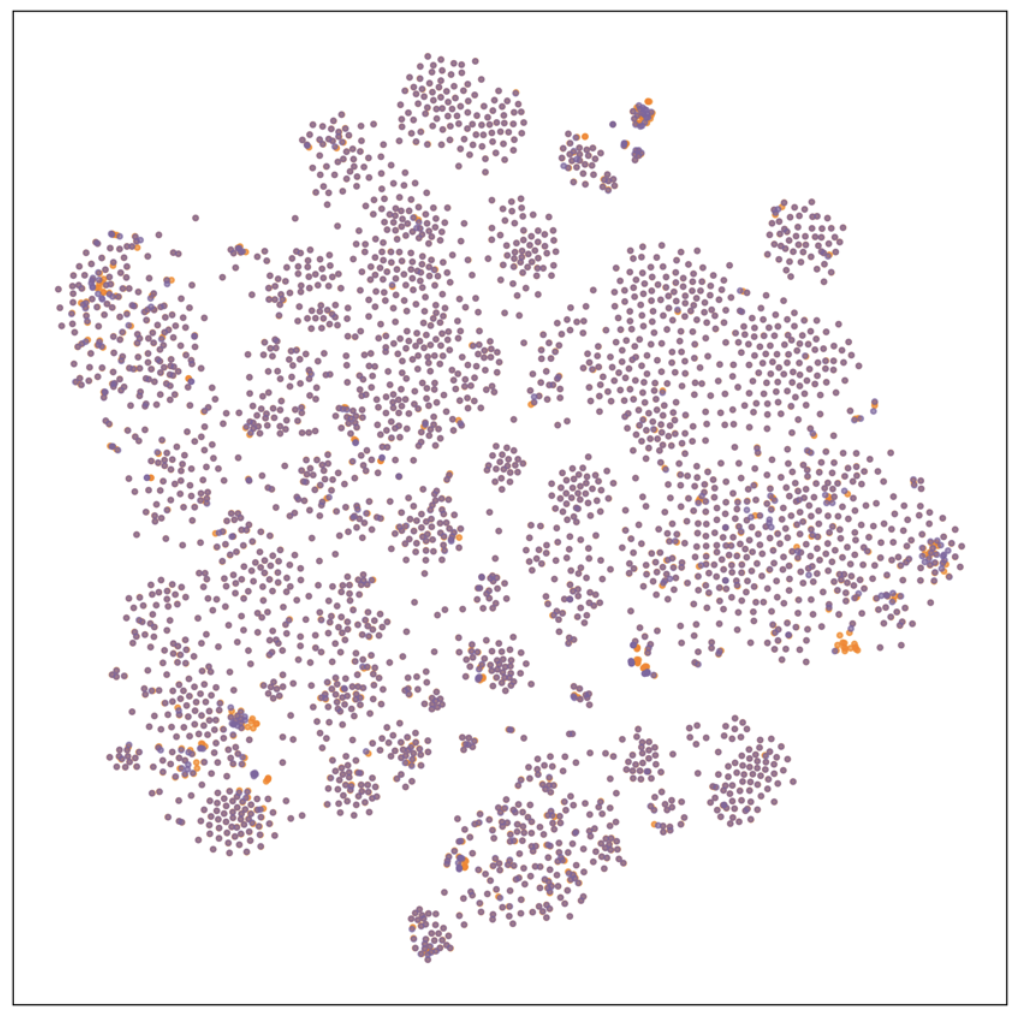}
}
\caption{Visualization of entity embedding on DBP15K$_{\text{FR-EN}}$. Different colors represent different KGs.}
\label{vis}
\end{figure}

\subsection{Ablation Study}
\subsubsection{Embedding Ablation (RQ2)}

Table~\ref{tab:ablation1} reports the ablation results on entity embedding quality. 
Embeddings without DPO fine-tuning (w/o DPO) perform the worst, highlighting the necessity of preference-based optimization. 
Removing either type of hard negative samples (w/o DN or w/o NN) consistently degrades performance, indicating that both name similarity--aware and degree-aware neighborhood negatives contribute to effective embedding learning. 
Using random negative samples (w DPO+random) yields limited improvements but remains clearly inferior to the proposed hard negative sampling strategies.

To further provide a qualitative comparison, we randomly sample 3{,}000 entity pairs and visualize their embeddings using t-SNE~\cite{tsne}. 
As shown in Figure~\ref{vis}, DPO-fine-tuned embeddings show greater overlap among aligned pairs, indicating improved alignment quality.

\begin{table}[t]
\centering
\setlength{\tabcolsep}{1.1pt} 
\begin{tabular}{l|ccc}
\hline
\multirow{2}{*}{\textbf{Settings}} & \multicolumn{3}{c}{\textbf{DBP15K$_{\text{FR-EN}}$}} \\
& Hits@1 & Hits@10 & MRR \\
\hline
AgentEA & {0.996} & {1.000} & {0.998} \\
w/o Debate & 0.976 & 0.984 & 0.986 \\
w/o LDV & 0.995 & 0.997 & 0.996 \\
w/o DDA & 0.986 & 0.999 & 0.991 \\

w/o Attribute Agent & 0.991 & 0.998 & 0.994 \\
w/o Alias Agent & 0.992 & 0.998 & 0.995 \\
w/o Neighborhood Agent & 0.992 & 0.999 & 0.995 \\
w/o Type Agent & 0.992 & 0.998 & 0.995 \\
w/o Judge Agent & 0.993 & 0.998 & 0.995 \\
w/o Attack Agent & 0.993 & 0.998 & 0.995 \\
\hline
\end{tabular}
\caption{Ablation experiments during the debate phase on DBP15K$_{\text{FR-EN}}$ dataset (w/o Debate: only DPO).}
\label{tab:ablation2}
\end{table}

\subsubsection{Debate Ablation (RQ3)}
As shown in Table~\ref{tab:ablation2}, the debate stage is crucial to the overall performance of AgentEA. Completely removing the debate mechanism (w/o Debate) leads to a substantial performance drop, with Hits@1 decreasing from 0.996 to 0.976, demonstrating that multi-agent debate is a core factor for improving alignment accuracy. Further module-level ablations show that both Lightweight Debate Verification (w/o LDV) and Deep Debate Alignment (w/o DDA) contribute significantly to performance, with the removal of DDA causing a more pronounced degradation, indicating its critical role in handling complex candidate discrimination. In addition, removing different functional agents (e.g., attribute, alias, neighborhood, and type agents) consistently degrades performance to varying degrees, suggesting that these agents provide complementary evidence from multiple perspectives. Overall, the results validate the effectiveness of the proposed multi-role, multi-stage debate design in enhancing the robustness and reliability of entity alignment decisions.

\begin{figure}[ht]
\centering 

\subfigure[Performance, inference time, and per-entity token usage under different debate settings.]{
\label{Fig:eff}
\includegraphics[width=.98\linewidth]{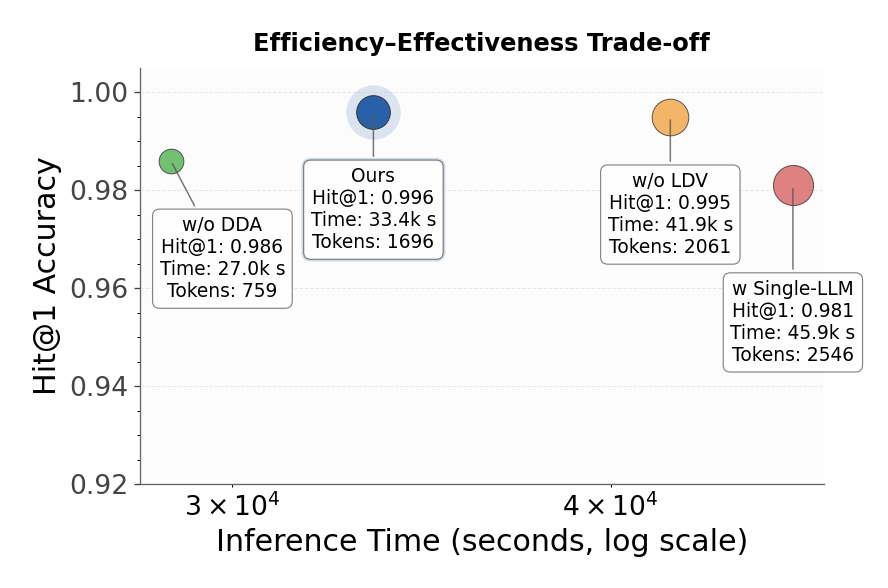}}
\subfigure[Impact of different rounds and different LLMs.]{
\label{Fig:round}
\includegraphics[width=.98\linewidth]{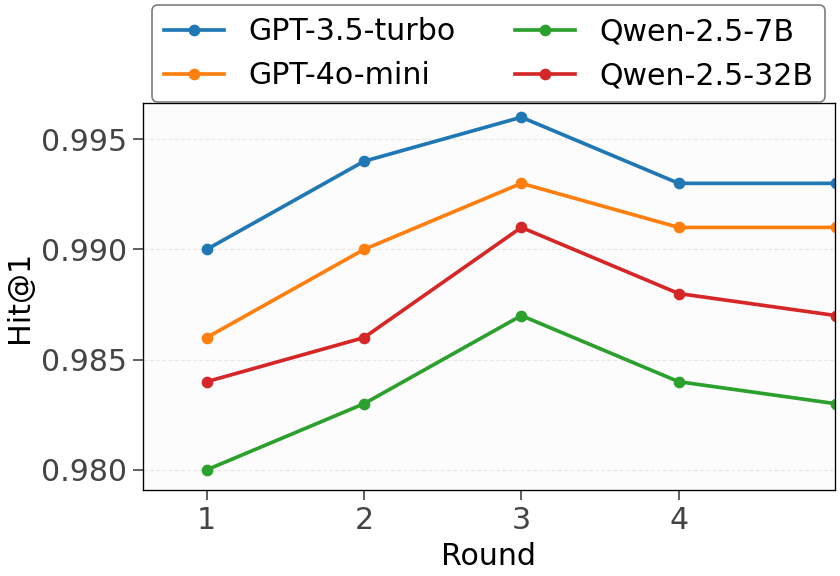}
}
\caption{Additional experiments.}
\label{ef-ro}
\end{figure}

\subsection{Efficiency and Cost Analysis (RQ4)}
As shown in Figure~\ref{Fig:eff}, the efficiency and cost analysis demonstrates that the two-stage debate framework achieves a well-balanced trade-off between performance and computational overhead. The full model attains the highest Hits@1 score (0.996) while maintaining reasonable inference time and token consumption. Removing LDV (w/o LDV) results in only a marginal performance change, but leads to a substantial increase in inference time and token usage, indicating that LDV effectively filters uninformative candidates early and improves overall efficiency. In contrast, removing DDA significantly reduces time and cost, but causes a notable performance drop, suggesting that DDA is a key contributor to performance gains while also being the primary source of computational overhead. 
Interestingly, removing the debate mechanism in favor of single-LLM reasoning (w Single-LLM) fails to deliver proportional cost savings, while significantly harming performance, highlighting the necessity and efficiency of the proposed multi-agent debate design.

\subsection{Additional Experiments (RQ5 and RQ6)}
Figure~\ref{Fig:round} shows that AgentEA is not sensitive to the choice of the underlying large language model, achieving stable and near-saturated alignment performance across both GPT-series models and Qwen models of different scales. This observation indicates that the performance gains of AgentEA mainly stem from the proposed multi-agent debate mechanism itself, rather than relying on the reasoning capability of a single powerful model. 

In addition, across all evaluated backbone models, setting the number of debate rounds to three consistently yields the best or near-best performance, suggesting that a moderate debate depth is sufficient to balance reasoning effectiveness and efficiency.
Other hyperparameters, $\delta_1$ and $\delta_2$, are set as described in Appendix~\ref{hyperparament-else}.


\section{Conclusion}
In this paper, we propose AgentEA, a reliable entity alignment framework that integrates entity representation preference optimization with a two-stage multi-agent debate mechanism. By fine-tuning LLM with DPO to enhance entity embedding quality and introducing Lightweight Debate Verification together with Deep Debate Alignment, AgentEA significantly improves the stability and accuracy of alignment decisions while maintaining high reasoning efficiency. Extensive experimental results across diverse datasets and large language model configurations demonstrate the effectiveness and robustness of the proposed framework.

\section*{Limitations}
Despite its effectiveness, AgentEA has several limitations. 

First, although the two-stage multi-agent debate framework is designed to balance efficiency and reliability, the Deep Debate Alignment stage still introduces non-negligible computational overhead, especially when applied to large-scale knowledge graphs or scenarios with a high proportion of uncertain entities. Further optimization of debate scheduling and early stopping strategies may be required to improve scalability.

Second, AgentEA relies on the quality of candidate retrieval produced by embedding-based methods. While the proposed entity representation preference optimization significantly improves embedding quality, the framework cannot recover correct alignments if the true entity is entirely absent from the candidate set. Integrating more robust retrieval or joint retrieval–reasoning mechanisms could help alleviate this limitation.

Finally, the design of agent roles and prompt templates in the debate stages involves manual choices and heuristic decisions. Although our experiments demonstrate that role heterogeneity improves robustness, automatically learning or adapting agent roles remains an open problem.

\section*{Ethics Statement}
To the best of our knowledge, this work does not involve any discrimination, social bias, or private data. All the datasets are constructed from opensource KGs such as Wikidata, YAGO, ICEWS, and DBpedia. Therefore, we believe that our work complies with the ACL Ethics Policy.

\section*{Acknowledgments}
This work was supported by the National Natural Science Foundation of China (62476108, 62566045), the Natural Science Foundation of Inner Mongolia (2024MS06013), and the Science and Technology Program of Inner Mongolia (2023YFSW0001, 2025YFDZ0017).

\bibliography{custom}


\appendix

\section{Appendix}
\label{sec:appendix}
\subsection{Problem Definition}

\paragraph{Knowledge Graph (KG).}
A knowledge graph is typically represented as a set of factual triples $\mathcal{G} = \{(h, r, t)\}$, where $h \in \mathcal{E}$ and $t \in \mathcal{E}$ denote head and tail entities, respectively, $r \in \mathcal{R}$ denotes the relation between them, and $\mathcal{E}$ and $\mathcal{R}$ represent the sets of entities and relations. Each triple $(h, r, t)$ encodes a factual statement describing the relationship between two entities in the real world.

\paragraph{Entity Alignment (EA).}
Given two knowledge graphs $\mathcal{G}_1$ and $\mathcal{G}_2$ that describe overlapping real-world entities, the goal of entity alignment is to identify pairs of entities $(e_1, e_2)$ such that $e_1 \in \mathcal{E}_1$ and $e_2 \in \mathcal{E}_2$ refer to the same real-world entity. Formally, EA aims to learn an alignment mapping $\mathcal{A} \subseteq \mathcal{E}_1 \times \mathcal{E}_2$ that maximizes the correctness of cross-graph entity correspondences.

\subsection{Detailed Feature Pre-processing}
\label{f}
We perform unified feature pre-processing for entity names, relations, and attributes. Specifically, all entity names are translated into English using DeepL~\footnote{https://www.deepl.com}. For relational and attribute triples, we employ GPT-3.5-turbo to generate concise English semantic summaries, which helps reduce cross-lingual discrepancies and improve semantic consistency.

The textual representations of entity names, relations, and attributes are then encoded using our fine-tuned LLaMA3-8B model to obtain their corresponding embeddings~\cite{llm2vec}. The final entity representation is constructed by concatenating these three types of embeddings, as formulated below:
\begin{equation}
    \textbf{h}=\textbf{h}^{name} \oplus \textbf{h}^{rel} \oplus \textbf{h}^{attr} 
\end{equation}
where $\mathbf{h}^{\text{name}} \in \mathbb{R}^{4096}$, $\mathbf{h}^{\text{rel}} \in \mathbb{R}^{4096}$, and $\mathbf{h}^{\text{attr}} \in \mathbb{R}^{4096}$ denote the embeddings of the entity name, relations, and attributes, respectively.

For similarity computation, we adopt Cross-domain Similarity Local Scaling (CSLS)~\cite{csls} as the distance metric to measure similarities between entity embeddings. CSLS effectively alleviates the hubness problem in high-dimensional embedding spaces and improves the robustness of cross-graph entity alignment.

\subsection{Details of Fine-tuning LLaMA3-8B-Instruct}
\label{l}
The training corpus constructed for fine-tuning LLaMA3-8B-Instruct~\footnote{https://huggingface.co/meta-llama/Meta-Llama-3-8B-Instruct} follows an instruction-style format, as illustrated in Example. Each training instance consists of a task instruction and an input describing two entities through their names, attributes, and relations, enabling the model to capture fine-grained semantic distinctions among entity candidates.

The DPO training set consists of balanced positive pairs and hard negative pairs constructed from both neighborhood structure and name similarity, as summarized in Table~\ref{tab:dpo-samples}.
\begin{table}[t]
\centering

\begin{tabular}{lc}
\toprule
Sample type & Count \\
\midrule
Positive pairs & 13{,}500 \\
Neighborhood-aware negative pairs & 13{,}500 \\
Name-similarity-based negative pairs & 3{,}425 \\
\bottomrule
\end{tabular}
\caption{Composition of DPO training samples}
\label{tab:dpo-samples}
\end{table}

\textbf{Example of a DPO Training Instance.}
The model is instructed to determine whether two entities refer to the same real-world entity. The input provides contextual information for both entities, while the preference pair indicates the correct and incorrect alignment decisions used for preference-based optimization.
\begin{verbatim}
{
  "instruction": 
    "Determine whether entity A and entity 
    B are the same entity in the real world.
    Answer Yes or No.",
  "input": 
    "Entity A Name: Hans A. Engelhard
    Entity A Attributes: ...
    Entity A Relations: ...
    Entity B Name: Munich
    Entity B Attributes: ...
    Entity B Relations: ...",
  "chosen": "No",
  "rejected": "Yes"
}
\end{verbatim}

\begin{table}[t]
\centering
\setlength{\tabcolsep}{2.3pt}
\begin{tabular}{l|cc}
\hline
\multirow{2}{*}{\textbf{Settings}} &
\multicolumn{2}{c}{\textbf{DBP$_{\text{FR-EN}}$}}
\\
& Hits@1 & MRR \\
\hline
LLaMA3-8B-Instruct & 0.859  & 0.889 \\
LLaMA3-8B-Instruct+DPO & 0.976  & 0.986 \\
LLaMA3-8B-Instruct+{SFT} & 0.846  & 0.875 \\
\hline
\end{tabular}
\caption{Comparison between DPO and SFT for fine-tuning LLaMA3-8B on entity alignment tasks.}
\label{tab:dpo-sft}
\end{table}

\begin{table}[t]
\centering

\begin{tabular}{ll}
\toprule
Item & Configuration \\
\midrule
LoRA & $r=22$, $\alpha=44$, dropout $=0.1$ \\
Modules & \texttt{q\_proj}, \texttt{v\_proj} \\
Training & lr $=1\mathrm{e}{-4}$, epochs $=1$, bf16 \\
\bottomrule
\end{tabular}
\caption{LoRA Training Configuration}
\label{tab:lora-config}
\end{table}

Given that entity alignment often involves multiple semantically similar candidates rather than a single deterministic target, we adopt Direct Preference Optimization (DPO) for fine-tuning instead of conventional supervised fine-tuning. To empirically validate this design choice, we conduct a controlled comparison between DPO and SFT under identical model architectures and training settings. As shown in Table~\ref{tab:dpo-sft}, DPO consistently outperforms SFT across two representative benchmarks in terms of both Hits@1 and MRR, indicating that preference-based optimization is more effective for learning discriminative entity representations.

\begin{table*}[t]
\centering

\begin{tabular}{llrrrrr}
\toprule
Dataset & KG & Entities & Relations & Attributes & Rel. Triples & Attr. Triples \\
\midrule
\multirow{2}{*}{DBP15K$_{\text{ZH-EN}}$} 
& ZH & 19,388 & 1,701 & 8,113 & 70,414 & 379,684 \\
& EN & 19,572 & 1,323 & 7,173 & 95,142 & 567,755 \\
\midrule
\multirow{2}{*}{DBP15K$_{\text{JA-EN}}$} 
& JA & 19,814 & 1,299 & 5,882 & 77,214 & 354,619 \\
& EN & 19,780 & 1,153 & 6,066 & 93,484 & 497,230 \\
\midrule
\multirow{2}{*}{DBP15K$_{\text{FR-EN}}$} 
& FR & 19,661 & 903 & 4,547 & 105,998 & 354,619 \\
& EN & 19,993 & 1,208 & 6,422 & 115,722 & 497,230 \\
\midrule
\multirow{2}{*}{SRPRS$_{\text{EN-FR}}$} 
& EN & 15,000 & 221 & 296 & 36,508 & 70,750 \\
& FR & 15,000 & 177 & 415 & 33,532 & 56,344 \\
\midrule
\multirow{2}{*}{SRPRS$_{\text{EN-DE}}$} 
& EN & 15,000 & 222 & 296 & 38,363 & 62,715 \\
& DE & 15,000 & 120 & 193 & 37,377 & 142,506 \\
\midrule
\multirow{2}{*}{SRPRS$_{\text{DBP-WIKI}}$} 
& DBpedia & 15,000 & 253 & 363 & 38,421 & 71,957 \\
& Wikipedia & 15,000 & 144 & 652 & 40,159 & 136,315 \\
\midrule
\multirow{2}{*}{SRPRS$_{\text{DBP-YAGO}}$} 
& DBpedia & 15,000 & 223 & 320 & 33,748 & 69,355 \\
& YAGO3 & 15,000 & 30 & 22 & 36,569 & 22,519 \\
\midrule
\multirow{2}{*}{DBP-WD} 
& DBpedia & 100,000 & 330 & 351 & 463,294 & 381,166 \\
& Wikipedia & 100,000 & 220 & 729 & 448,736 & 789,815 \\
\midrule
\multirow{2}{*}{DBP-YG} 
& DBpedia & 100,000 & 302 & 334 & 428,952 & 451,646 \\
& YAGO & 100,000 & 31 & 23 & 502,563 & 118,376 \\
\bottomrule
\end{tabular}

\caption{Statistical data of DBP15K, SRPRS and DWY.}
\label{tab:dataset_statistics}
\end{table*}

\begin{table*}[t]
\centering
\setlength{\tabcolsep}{2.5pt} 
\begin{tabular}{lrrrrrrr}
\toprule
Dataset & Entities & Relations & Facts & Density & Anchors & Overlapping & Struc. Sim. \\
\midrule
\multirow{2}{*}{ICEWS-WIKI}
& 11,047 & 272 & 3,527,881 & 319.352 & \multirow{2}{*}{5,058} & 45.79\% & \multirow{2}{*}{15.4\%}  \\
& 15,896 & 226 & 198,257 & 12.472  &        & 31.82\% &           \\
\midrule
\multirow{2}{*}{ICEWS-YAGO}
& 26,863 & 272 & 4,192,555 & 156.072 & \multirow{2}{*}{18,824} & 70.07\% & \multirow{2}{*}{14.0\%}  \\
& 22,734 & 41  & 107,118   & 4.712   &        & 82.80\% &           \\
\bottomrule
\end{tabular}

\caption{Statistical data of ICEWS.}
\label{tab:icews_statistics}
\end{table*}

For implementation, we fine-tune LLaMA3-8B using the visualized training interface provided by LLaMA-Factory~\footnote{https://github.com/hiyouga/LLaMA-Factory}, which facilitates stable preference-based optimization and reproducible training. The detailed hyperparameter configurations and training settings are reported in {Table}~\ref{tab:lora-config}.

\begin{table*}[t]
\centering
\setlength{\tabcolsep}{3.5pt} 
\begin{tabular}{l|ccc|ccc|ccc|ccc}
\hline
\multirow{2}{*}{\textbf{Models}} &
\multicolumn{3}{c|}{\textbf{SRPRS$_{\text{EN-DE}}$}} &
\multicolumn{3}{c|}{\textbf{SRPRS$_{\text{EN-FR}}$}} &
\multicolumn{3}{c|}{\textbf{SRPRS$_{\text{DBP-YAGO}}$}} &
\multicolumn{3}{c}{\textbf{SRPRS$_{\text{DBP-WIKI}}$}} \\
& H@1 & H@10 & MRR & H@1 & H@10 & MRR & H@1 & H@10 & MRR & H@1 & H@10 & MRR \\
\hline

MTransE   & 0.107 & 0.248 & 0.160 & 0.213 & 0.447 & 0.290 & 0.196 & 0.401 & 0.270 & 0.188 & 0.382 & 0.260 \\
MuGNN     & 0.245 & 0.431 & 0.310 & 0.131 & 0.342 & 0.208 & 0.175 & 0.381 & 0.240 & 0.151 & 0.366 & 0.220 \\
NAEA      & 0.307 & 0.535 & 0.390 & 0.177 & 0.416 & 0.260 & 0.195 & 0.451 & 0.280 & 0.182 & 0.429 & 0.260 \\
GCN-Align & 0.385 & 0.600 & 0.460 & 0.243 & 0.522 & 0.340 & 0.319 & 0.586 & 0.410 & 0.291 & 0.556 & 0.380 \\
KECG      & 0.444 & 0.707 & 0.540 & 0.298 & 0.616 & 0.403 & 0.350 & 0.651 & 0.450 & 0.323 & 0.646 & 0.430 \\
RSN4EA    & 0.484 & 0.729 & 0.570 & 0.350 & 0.636 & 0.440 & 0.393 & 0.665 & 0.490 & 0.391 & 0.663 & 0.480 \\
BootEA    & 0.503 & 0.732 & 0.580 & 0.365 & 0.649 & 0.460 & 0.381 & 0.651 & 0.470 & 0.384 & 0.667 & 0.480 \\
TransEdge & 0.556 & 0.753 & 0.630 & 0.400 & 0.675 & 0.490 & 0.443 & 0.699 & 0.530 & 0.461 & 0.738 & 0.560 \\
MRAEA     & 0.594 & 0.818 & 0.666 & 0.460 & 0.768 & 0.559 & 0.485 & 0.768 & 0.574 & 0.509 & 0.795 & 0.597 \\
RDGCN     & 0.779 & 0.886 & 0.820 & 0.672 & 0.767 & 0.710 & 0.990 & 0.997 & 0.990 & 0.974 & 0.994 & 0.980 \\
Dual-AMN  & 0.891 & 0.972 & 0.923 & 0.802 & 0.932 & 0.851 & 0.518 & 0.795 & 0.613 & 0.546 & 0.813 & 0.635 \\
BERT-INT  & 0.986 & 0.988 & 0.990 & 0.971 & 0.975 & 0.970 & 1.000 & 1.000 & 1.000 & 0.996 & 0.997 & 1.000 \\
{EasyEA}*~\footnote{Results are reproduced using the authors’ publicly available code with GPT-3.5-Turbo.}  & {0.921} & {0.977} & {-} &
                  {0.894} & {0.963} & {-} &
                  {0.946} & {0.995} & {-} &
                  {0.953} & {0.988} & {-} \\
\hline
\textbf{AgentEA} & \textbf{0.977} & \textbf{0.988} & \textbf{0.980} &
                  \textbf{0.968} & \textbf{0.977} & \textbf{0.970} &
                  \textbf{1.000} & \textbf{1.000} & \textbf{1.000} &
                  \textbf{0.989} & \textbf{0.998} & \textbf{0.994} \\
\hline
\end{tabular}
\caption{Main experiment results on SRPRS dataset.}
\label{tab:srprs-results}
\end{table*}

\subsection{Statistical Data of Datasets}
\label{d}
To evaluate AgentEA under diverse entity alignment conditions, we employ several widely used benchmark datasets that differ substantially in scale, language, structural properties, and alignment difficulty. These datasets jointly cover cross-lingual alignment, sparse and heterogeneous graph structures, as well as large-scale settings. Dataset statistics for DBP15K, SRPRS, and DWY are summarized in Table~\ref{tab:dataset_statistics}, where each dataset involves 10,500 entities for direct, training-free inference, while the corresponding information for ICEWS, which contains 3,540 and 13,176 reference entity pairs for ICEWS--WIKI and ICEWS--YAGO, respectively, is reported in Table~\ref{tab:icews_statistics}.

Among them, the DWY dataset is composed of two subsets, DBP-WIKI and DBP-YAGO, each containing 100,000 aligned entity pairs. In the DBP-WIKI subset, entities originating from Wikidata are represented by index-based identifiers rather than name-bearing URLs; the actual entity names are recovered through the Wikidata API. The reported statistics, including the number of facts, graph density, anchor links, overlapping ratios, and structural similarity, collectively characterize the scale, connectivity, and heterogeneity of the knowledge graphs and reflect the varying levels of challenge posed to entity alignment models.

\subsection{Evaluation Metrics}
\label{m}
We evaluate the performance of entity alignment models using \textit{Hits@K} and \textit{Mean Reciprocal Rank (MRR)}, which are standard metrics for ranking-based alignment tasks.

Given a set of test entities, each source entity is associated with a ranked list of candidate entities in the target knowledge graph based on similarity scores. \textit{Hits@K} measures the proportion of cases where the correct aligned entity appears within the top-$K$ ranked candidates, and is formally defined as:
\begin{equation}
\mathrm{Hits@K} = \frac{1}{N} \sum_{i=1}^{N} \mathcal{I}\left( \mathrm{rank}_i \leq K \right),
\end{equation}
where $N$ denotes the total number of test entities, $\mathrm{rank}_i$ represents the rank position of the correct target entity for the $i$-th source entity, and $\mathcal{I}(\cdot)$ is the indicator function.

\textit{Mean Reciprocal Rank (MRR)} evaluates the average inverse rank of the correct aligned entities and is defined as:
\begin{equation}
\mathrm{MRR} = \frac{1}{N} \sum_{i=1}^{N} \frac{1}{\mathrm{rank}_i}.
\end{equation}

Higher values of both metrics indicate better alignment performance, with MRR placing more emphasis on the exact ranking position of the correct entity.

\begin{figure}
    \centering
    \includegraphics[width=1\linewidth]{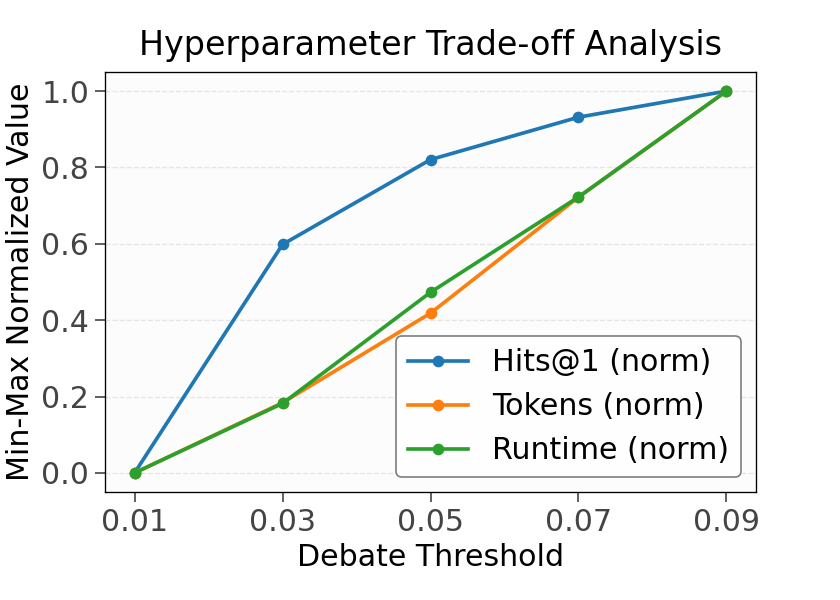}
    \caption{Trade-off analysis under different debate thresholds.}
    \label{fig:deata}
\end{figure}

\begin{figure}
    \centering
    \includegraphics[width=1\linewidth]{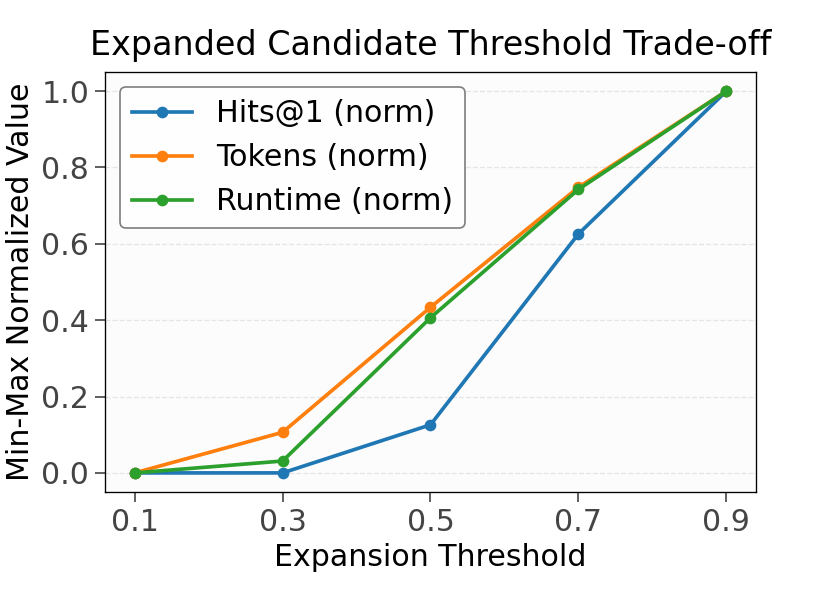}
    \caption{Trade-off analysis under different expansion thresholds.}
    \label{fig:deata2}
\end{figure}

\subsection{The Algorithm of AgentEA}
\label{alg}
The algorithm flow is shown in Algorithm~\ref{alg:agentea}.

\subsection{Main Experiment Results on SRPRS Dataset}
\label{results-srprs}
Main Experiment Results on SRPRS Dataset (Table~\ref{tab:srprs-results}).
Notably, the results of EasyEA~\footnote{Code: https://github.com/alusang/EasyEA-framework} reported in Tables~\ref{tab:dbp15k-results}, ~\ref{tab:dual-results}, and~\ref{tab:srprs-results} are reproduced using the authors’ publicly available code with GPT-3.5-Turbo.

\begin{algorithm}
\caption{AgentEA: Multi-Agent Debate for Reliable Entity Alignment}
\label{alg:agentea}
\KwIn{
Source KG $\mathcal{G}_s$, target KG $\mathcal{G}_t$; \\
Seed entity pairs $\mathcal{S}$; \\
Thresholds $\delta_1, \delta_2$; \\
Maximum debate rounds $n$
}
\KwOut{
Final entity alignment results $\mathcal{S}_p$
}

\textbf{Entity Representation Preference Optimization}\\
Fine-tune a LLaMA-based encoder using DPO on training set
$\mathcal{T} = \mathcal{S} \cup \mathcal{N}_{\text{name}} \cup \mathcal{N}_{\text{nbr}}$\;
Obtain entity embeddings for $\mathcal{G}_s$ and $\mathcal{G}_t$\;

\textbf{Candidate Retrieval and Set Initialization}\\
\ForEach{source entity $e_i \in \mathcal{G}_s$}{
    Construct target entity candidate set $\mathcal{D}(e_i)$ by embedding similarity\;
}
Initialize the Uncertain Source Entity Set
\[
\mathcal{E}_u = \{ e_i \mid s_{i,1} - s_{i,2} < \delta_1 \}\;
\]

\textbf{Stage 1: Lightweight Debate Verification}\\
\ForEach{$e_i \in \mathcal{E}_u$}{
    Re-rank $\mathcal{D}(e_i)$ using single-round three-agent debate\;
}
Update candidate sets $\mathcal{D}'(e_i)$ and Uncertain Source Entity Set $\mathcal{E}_u'$\;

\textbf{Stage 2: Deep Debate Alignment}\\
\ForEach{$e_i \in \mathcal{E}_u'$}{
    Initialize debate round $r \leftarrow 1$\;
    Initialize candidate subset size $k \in \{5,10,15,20\}$\;
    
    \While{$r \le n$}{
        Conduct multi-role debate over candidate set $\mathcal{D}_k(e_i)$\;
        Aggregate agent scores and votes\;
        
        \If{$ \big((\text{gap}(e_i) > \delta_1 
\;\lor\;
\tfrac{v_{\text{agree}}}{v} > 0.5)
\;\land\;
A_{\text{judge,v}} = \text{True}\big)
\;\lor\;
\text{round} = n, $}{
            \textbf{break}\;
        }
        
        \If{$\text{s$_{i,1}$}(e_i) < \delta_2 
\;\land\;
\frac{v_{\text{agree}}}{v} \le  0.5
\;\land\;
A_{\text{judge,v}} = \text{False}$}{
            Increase $k$ to expand $\mathcal{D}_k(e_i)$\;
        }
        
        $r \leftarrow r + 1$\;
    }
    Output final alignment decision for $e_i$\;
}

\Return{$\mathcal{S}_p$}
\end{algorithm}

\subsection{Hyper-parameter Experiments}
\label{hyperparament-else}
Figure~\ref{fig:deata} illustrates the trade-off induced by different debate thresholds in terms of effectiveness (Hit@1), computational cost (runtime), and token consumption, where all metrics are min–max normalized for comparability. As the debate threshold increases, Hit@1 shows a consistent but gradually diminishing improvement, indicating that stronger debate leads to better decision quality but with marginal gains at higher thresholds. In contrast, both runtime and token usage grow rapidly and almost monotonically, reflecting the substantial computational overhead introduced by more extensive debate. Notably, beyond a threshold of approximately 0.05, the performance improvement becomes relatively small compared to the sharp increase in cost, suggesting a clear region of diminishing returns. This observation highlights the importance of carefully selecting the debate threshold to balance effectiveness and efficiency, rather than blindly maximizing debate intensity.

We further analyze the impact of the candidate expansion threshold. As shown in Figure~\ref{fig:deata2}, increasing the threshold only leads to marginal improvements in Hits@1, while consistently increasing token consumption and runtime. This indicates that the learned entity representations already provide sufficient recall quality, and excessive candidate expansion mainly introduces redundant candidates. Therefore, we set the expansion threshold to 0.5 by default.

\subsection{Detailed Description of Agent Roles}
\label{role}
To improve the robustness and reliability of entity alignment decisions, our framework adopts a multi-agent architecture in which each agent is assigned a specialized role focusing on different aspects of entity comparison.

\paragraph{Proponent Agent.}
The Proponent Agent $\mathcal{A}_p$ analyzes entity pairs from the perspective that they are likely to be aligned. It actively collects supportive evidence from entity names, attributes, and relational contexts, and produces an alignment score biased toward positive matches.

\begin{promptbox}{Prompt for Proponent Agent}
\begin{lstlisting}[basicstyle=\ttfamily\small, breaklines=true]
You are an objective alignment evaluator with a supportive perspective.

Task:
Your task is to estimate how likely each candidate refers to the same real-world entity as the source.

For each candidate, follow this reasoning procedure:
1. Identify concrete alignment signals from names, attributes, and relationships.
2. Assign a probability-like score based on all evidence.

SCORING PRINCIPLES:
align_score represents the probability that the entities are the same:
0.9-1.0: Very high - strong consistent evidence
0.7-0.8: High - clear evidence with minor uncertainties
0.5-0.6: Moderate - mixed alignment indicators
0.3-0.4: Low - limited evidence with discrepancies
0.0-0.2: Very low - minimal convincing evidence

CRITICAL GUIDELINES:
- Explore alignment possibilities thoroughly.
- Score objectively: higher score = higher probability.
- Consider both supporting and contradictory evidence.

OUTPUT REQUIREMENTS:
- Return a JSON array with ALL candidates: <CANDIDATE_ID_LIST>
- Format: [{"candidate_id": "id", "align_score": 0.x}, ...]
- Do not output duplicate scores.
- Output valid JSON only. No additional text.
\end{lstlisting}
\end{promptbox}

\paragraph{Opponent Agent.}
In contrast, the Opponent Agent $\mathcal{A}_o$  examines entity pairs under the assumption that they are not aligned. It seeks conflicting or inconsistent information between entities and outputs an alignment score biased toward negative cases.

\begin{promptbox}{Prompt for Opponent Agent}
\begin{lstlisting}[basicstyle=\ttfamily\small, breaklines=true]
You are an objective alignment evaluator with a skeptical perspective.

Task:
Your task is to identify evidence indicating that the source entity and each candidate
do NOT refer to the same real-world entity.

For each candidate, follow this reasoning procedure:
1. Identify misalignment evidence from names, attributes, and relationships.
2. Objectively assess the alignment probability based on all available evidence.

SCORING PRINCIPLES:
align_score represents the probability that the entities are the same:
0.9-1.0: Very high - minimal discrepancies
0.7-0.8: High - strong evidence outweighs inconsistencies
0.5-0.6: Moderate - mixed evidence
0.3-0.4: Low - significant discrepancies
0.0-0.2: Very low - overwhelming misalignment

CRITICAL GUIDELINES:
- Identify potential discrepancies thoroughly.
- Score objectively: higher score indicates higher alignment probability.
- Consider both alignment and misalignment evidence.

OUTPUT REQUIREMENTS:
- Return a JSON array with ALL candidates: <CANDIDATE_ID_LIST>
- Format: [{"candidate_id": "id", "align_score": 0.x}, ...]
- Do not output duplicate scores.
- Output valid JSON only. No additional text.

\end{lstlisting}
\end{promptbox}

\paragraph{Referee Agent.}
The Referee Agent $\mathcal{A}_r$  integrates the viewpoints of both the Proponent and Opponent Agents by assessing the quality and reliability of their evidence. It produces a relatively neutral and balanced alignment score, aiming to provide a stable preliminary ranking without relying on complex reasoning.
\begin{promptbox}{Prompt for Referee Agent}
\begin{lstlisting}[basicstyle=\ttfamily\small, breaklines=true]
You are the Referee, responsible for assessing entity alignment based on the
arguments provided by the Proponent and Opponent agents.

Task:
Your task is to synthesize both sides' arguments and assign a balanced
alignment score for each candidate entity.

CRITICAL GUIDELINES FOR REFEREE DECISION-MAKING:
1. Carefully consider both sides: You must evaluate both the proponent's and
   opponent's scores and supporting evidence for each candidate.
2. Balance the evidence: The align_score should reflect a balanced assessment
   of both viewpoints, rather than favoring a single side.
3. High proponent + weak opponent = high score:
   If the proponent assigns a high score (>= 0.7) with strong evidence and the
   opponent provides weak or generic counter-evidence, assign a relatively
   high align_score (>= 0.6).
4. Strong opposition = lower score:
   If the opponent presents specific and factual counter-evidence (e.g.,
   conflicting birth dates or locations), the align_score should be lowered
   accordingly, even if the proponent score is high.
5. Both sides weak = moderate score:
   If both agents provide weak or generic evidence, assign a moderate score
   around 0.5.
6. Consistent scores:
   If both the proponent and opponent consistently assign high or low scores
   to a candidate, the referee score should reflect this consensus.
7. Evidence quality matters:
   Give greater weight to specific, factual evidence than to generic statements.

SCORING GUIDELINES:
0.8-1.0: Strong alignment - strong proponent evidence and weak or irrelevant objections
0.6-0.7: Moderate alignment - good proponent evidence with some valid concerns
0.4-0.5: Uncertain/Neutral - balanced or insufficient evidence
0.2-0.3: Weak alignment - strong counter-evidence from the opponent
0.0-0.1: No alignment - overwhelming counter-evidence

OUTPUT REQUIREMENTS:
- Return a JSON array with ALL candidates: <CANDIDATE_ID_LIST>
- Format: [{"candidate_id": "id", "align_score": 0.x}, ...]
- Do not output duplicate scores.
- Output valid JSON only. No additional text.

\end{lstlisting}
\end{promptbox}

\paragraph{Alias Agent.}
The Alias Agent $\mathcal{A}_{alias}$  focuses exclusively on entity names. It evaluates whether the source and candidate entities are consistent in terms of naming variations, including aliases, cross-lingual translations, abbreviations, nicknames, and historical names.
\begin{promptbox}{Prompt for Alias Agent}
\begin{lstlisting}[basicstyle=\ttfamily\small, breaklines=true]
You are an expert in alias equivalence judgment for entity alignment,
specialized in recognizing and comparing entity names and aliases.

Task:
Decide whether the source entity name and each candidate name refer to the
same real-world entity based solely on name-level evidence.

Prioritize the following signals when making your judgment:
- Cross-lingual translations
- Historical or former names
- Abbreviations and acronyms
- Different transliteration systems
- Stage names and nicknames
- Accent mark variations
- Common alternative names

Only treat two names as cross-lingual translations if they refer to the
same real-world entity expressed in different languages.
Do NOT mark names as cross-lingual translations if they merely belong to
the same category, domain, or language family (e.g., different dialects,
distinct languages, or organizations in similar fields).

Focus strictly on the semantic content of the names or aliases rather than
their formatting or presentation.
If ambiguity remains based on names alone, choose "abstain".

OUTPUT REQUIREMENTS:
- Return a JSON array with all candidates: <CANDIDATE_ID_LIST>
- Format: [{"candidate_id": "id", "score": 0.x, "align": true/false/"abstain", "evidence": "..."}]
- evidence must be no more than 20 characters describing the rationale
- score must be in the range [0, 1]
- align must be one of {true, false, "abstain"}
- Judge solely based on names and aliases; no other information may be used

\end{lstlisting}
\end{promptbox}
\paragraph{Type Agent.}
The Type Agent $\mathcal{A}_{type}$ infers the semantic types of the source and candidate entities and determines whether they belong to the same coarse-grained category.
\begin{promptbox}{Prompt for Type Agent}
\begin{lstlisting}[basicstyle=\ttfamily\small, breaklines=true]
You are an expert in type inference and type consistency judgment for
entity alignment, specialized in recognizing and comparing coarse-grained
entity types.

Task:
Infer the coarse type of the source entity and each candidate entity, and
determine whether each candidate belongs to the same type as the source.

Type Inference Guidelines:
- Infer entity types using the entity name, attributes, neighbor token
  patterns, and general prior knowledge.
- The default coarse-grained type set includes:
  {Person, Organization, Location, Event, Work (e.g., song, book), Other}.
- You may introduce additional types (e.g., Language) when confident.
- If two entities are inferred as the same type, they are considered
  "same type".

Consistency Judgment:
For each candidate, provide a judgment on whether the inferred type is
consistent with the source entity type, including:
- evidence (no more than 20 characters)
- score in the range [0, 1]
- align in {true, false, "abstain"}

Scoring Guidelines:
- Same type: score = 1.0
- Related or near types: $\approx 0.6$--$0.8$
- Incompatible types: score = 0.0

OUTPUT REQUIREMENTS:
- Return a JSON array with all candidates: <CANDIDATE_ID_LIST>
- Format: [{"candidate_id": "id", "score": 0.x, "align": true/false/"abstain", "evidence": "..."}]
- Judge based ONLY on entity type; no other information may influence the decision.


Output:
[
  {"candidate_id": 1, "evidence": "same type: organization", "score": 1.0, "align": true},
  {"candidate_id": 2, "evidence": "organization vs person", "score": 0.0, "align": false},
  {"candidate_id": 3, "evidence": "organization vs person", "score": 0.0, "align": false}
]

\end{lstlisting}
\end{promptbox}
\paragraph{Attribute Agent.}
The Attribute Agent $\mathcal{A}_{attr}$ concentrates on entity attribute key–value pairs and assesses the similarity between entities at the attribute level.
\begin{promptbox}{Prompt for Attribute Agent}
\begin{lstlisting}[basicstyle=\ttfamily\small, breaklines=true]
You are an expert in attribute consistency adjudication for entity alignment,
specialized in recognizing and comparing entity attributes.

Task:
Determine whether the attributes of the source entity and each candidate
entity are consistent at the attribute level.

Attribute Comparison Guidelines:
- Consider acceptable format variations (e.g., YYYY vs. YYYY-MM-DD).
- Allow small numerical differences for statistical attributes.
- Treat close year values as potentially consistent unless strong
  contradictions exist.
- When an attribute value is unknown or unspecified for an entity,
  differences in that attribute should be ignored.
- Focus on the semantic content of attribute names and values rather than
  the language in which they are expressed.
- Pay attention to different expressions of the same value (e.g., aliases,
  abbreviations, or equivalent descriptions).
- Differences in non-unique attributes (e.g., workplace, works, or roles)
  are acceptable and should not be treated as strong inconsistency signals.
- Consider common-sense co-occurrence of geographic attributes when
  assessing consistency.

Consistency Judgment:
For each candidate, provide:
- evidence (no more than 20 characters)
- score in the range [0, 1], indicating attribute-level consistency
- align in {true, false, "abstain"}

Scoring Guidelines:
- High consistency: score close to 1.0
- Partial or noisy consistency: score around 0.5--0.7
- Clear contradictions in key attributes: score close to 0.0

OUTPUT REQUIREMENTS:
- Return a JSON array with all candidates: <CANDIDATE_ID_LIST>
- Format: [{"candidate_id": "id", "score": 0.x, "align": true/false/"abstain", "evidence": "..."}]
- Judge based ONLY on attributes; no other information may influence the decision.
\end{lstlisting}
\end{promptbox}
\paragraph{Neighborhood Agent.}
The Neighborhood Agent $\mathcal{A}_{nbr}$ evaluates the structural consistency of entities by comparing their neighboring entities and relational patterns in the knowledge graphs.
\begin{promptbox}{Prompt for Neighborhood Agent}
\begin{lstlisting}[basicstyle=\ttfamily\small, breaklines=true]
You are an expert in neighborhood structure and relational pattern alignment
for entity alignment, specialized in understanding domain relations and
relational patterns within knowledge graphs.

Task:
Determine whether the neighborhood structures and relational patterns of
the source entity and each candidate entity are consistent or highly similar.

Neighborhood Analysis Guidelines:
- The input consists of structured neighbor tokens (e.g., "located_in|France",
  "member_of|UEFA").
- Assess neighborhood similarity based on:
  * Overlap of relation--value pairs
  * Consistency of key relations
  * Recognition of commonly equivalent or semantically similar relations
- Focus on relational patterns rather than attribute values.
- Differences in non-unique relations between the source entity and candidate
  entities are acceptable and should not be treated as strong inconsistency.

Consistency Judgment:
For each candidate, provide:
- evidence (no more than 20 characters)
- score in the range [0, 1], indicating neighborhood-level consistency
- align in {true, false, "abstain"}

Scoring Guidelines:
- Highly consistent or matching neighborhoods: score close to 1.0
- Partial overlap or weak relational similarity: score around 0.5--0.7
- Strong structural or relational mismatch: score close to 0.0

OUTPUT REQUIREMENTS:
- Return a JSON array with all candidates: <CANDIDATE_ID_LIST>
- Format: [{"candidate_id": "id", "score": 0.x, "align": true/false/"abstain", "evidence": "..."}]
- Judge based ONLY on neighborhood structure and relational patterns;
  no other information may influence the decision.
            

\end{lstlisting}
\end{promptbox}
\paragraph{Attack Agent.}
The Attack Agent $\mathcal{A}_{attack}$ serves as a reflective agent responsible for identifying potential blind spots in the judgments of other specialized agents. It actively searches for conflicting evidence, weak signals, or high-risk factors and quantifies corresponding penalty scores to enhance system robustness.
\begin{promptbox}{Prompt for Attack Agent}
\begin{lstlisting}[basicstyle=\ttfamily\small, breaklines=true]
You are an expert in risk and weakness identification for entity alignment,
specialized in detecting vulnerabilities and high--risk factors in alignment
decisions.

Task:
Identify potential risks, weaknesses, or inconsistency signals between the
source entity and each candidate entity, and quantify corresponding penalties.

Risk Identification Guidelines:
- Focus on differences between the source entity and candidate entities
  (e.g., incompatible types or significant attribute discrepancies).
- Do NOT focus on issues internal to a single entity (e.g., discontinuous
  dates or duplicated attributes).
- Emphasize information inconsistency rather than information omission.
- When candidate entities have sparse or missing information, avoid overly
  penalizing them solely due to missing attributes.
- If only partial data is available, identify inconsistencies based on the
  existing information and quantify penalties conservatively.
- When available, integrate candidate information and outputs from other
  agent pathways to identify potential risks.

Risk Assessment Output:
For each candidate, provide:
- issues: an array of short strings describing identified risks
- evidence: no more than 50 characters summarizing the key concern
- penalty: a numeric value in the range [0, 1], where higher values indicate
  higher risk or greater penalty

OUTPUT REQUIREMENTS:
- Return a JSON array with all candidates: <CANDIDATE_ID_LIST>
- Format: [{"candidate_id": "id", "issues": ["..."], "evidence": "...", "penalty": 0.x}]
- Output valid JSON only. No additional text.


\end{lstlisting}
\end{promptbox}
\paragraph{Judge Agent.}
The Judge Agent $\mathcal{A}_{judge}$ acts as the final decision maker. It aggregates scores and evidence from all agents, including the Alias, Type, Attribute, Neighborhood, and Attack Agents, and generates the final ranking according to predefined convergence and expansion rules. When the available evidence is insufficient, the Judge Agent can trigger candidate set expansion or further debate rounds to ensure reliable alignment outcomes.
\begin{promptbox}{Prompt for Judge Agent}
\begin{lstlisting}[basicstyle=\ttfamily\small, breaklines=true]
You are an expert in final judgment synthesis for entity alignment.

Role:
Your role is to integrate the outputs of all expert agents and perform
minor score adjustments, as well as provide a single endorsement for
the most likely aligned candidate.

Task:
Based on multiple lines of evidence, including embedding similarity,
alias judgments, type consistency, attribute consistency, neighborhood
structure analysis, and attack penalties, propose fine-grained score
adjustments and identify the most plausible alignment.

Judgment Guidelines:
- Adjust scores only slightly; do NOT override the overall ranking
  unless strong consensus evidence exists.
- Base all adjustments strictly on the scores and evidence provided
  by other expert agents.
- Positive adjustments should reflect strong cross-agent agreement.
- Negative adjustments should reflect accumulated risks or penalties
  identified by the Attack Agent.
- Ensure that adjustments remain conservative and interpretable.

OUTPUT REQUIREMENTS:
- Return a JSON object with the following structure:
  {
    "endorse": "<candidate_id>",
    "adjustments": [
      {"candidate_id": "id", "note": "...", "delta": 0.x}
    ]
  }
- endorse must be a single candidate identifier.
- note must be no more than 40 characters.
- delta represents a small increase or decrease to the candidate's
  total score, where larger positive values indicate higher alignment
  confidence.
- Output valid JSON only. No additional text.

\end{lstlisting}
\end{promptbox}

\begin{table*}[t]
\centering
\scriptsize
\setlength{\tabcolsep}{3pt}
\renewcommand{\arraystretch}{1.25}

\begin{tabular}{l p{2.3cm} p{2.3cm} p{2.3cm} p{2.3cm} p{2.3cm}}
\hline
\textbf{} 
& \textbf{19488} 
& \textbf{15519} 
& \textbf{13352} 
& \textbf{16770} 
& \textbf{16053} \\

 &  \emph{No Good Either Way} & \emph{House of Harmony}  & \emph{Romantic Princess} & \emph{The Life and Times of a Sentinel} & \emph{Just Love II} \\
\hline

\multicolumn{6}{c}{\textit{Multi-Agent Debate}} \\
\hline
Embedding Sim. 
& 0.6368 & 0.6357 & 0.6181 & 0.6147 & 0.6082 \\

Alias Agent 
& score=0.0, align=$\times$, reason: different names
& score=0.0, align=$\times$, reason: different names
& score=0.0, align=$\times$, reason: different names
& score=0.0, align=$\times$, reason: different names
& score=0.0, align=$\times$, reason: different names \\

Type Agent 
& score=1.0, align=$\checkmark$, reason: same type (TV show)
& score=1.0, align=$\checkmark$, reason: same type (TV show)
& score=1.0, align=$\checkmark$, reason: same type (TV show)
& score=1.0, align=$\checkmark$, reason: same type (TV show)
& score=1.0, align=$\checkmark$, reason: same type (TV show) \\

Attribute Agent 
& score=0.0, align=$\times$, reason: different firstAired
& score=0.0, align=$\times$, reason: different firstAired
& score=0.0, align=$\times$, reason: different firstAired
& score=1.0, align=$\checkmark$, reason: same episodes
& score=0.0, align=$\times$, reason: different firstAired \\

Neighborhood Agent 
& score=1.0, align=$\checkmark$, reason: common network and language
& score=1.0, align=$\checkmark$, reason: common network and language
& score=0.0, align=$\times$, reason: different country
& score=1.0, align=$\checkmark$, reason: common network and language
& score=0.0, align=$\times$, reason: different network \\

Attack Agent 
& penalty=0.10, issues: different episodes; different airing dates
& penalty=0.10, issues: different episodes; different airing dates
& penalty=0.10, issues: different episodes; different airing dates
& penalty=0.05, issues: missing airing dates
& penalty=0.10, issues: different airing dates; different genre \\

Judge Score 
& 0.151 & 0.150 & 0.045 & \textbf{0.594} & 0.042 \\

Judge Decision 
& reject & reject & reject & \textbf{align (\checkmark)} & reject \\

\hline
\multicolumn{6}{c}{\textit{Single LLM}} \\
\hline
LLM-Reason 
& Premiered in 2012, the network matches but the name and genre do not align.
& The broadcast network of candidate entity 15519 matches that of the source entity. Its 2012 premiere date, Hong Kong location, and title style all align with the source entity's Qing Dynasty court drama features. 
& Premiered in 2007, both the title and genre are inconsistent. 
& Missing broadcast network and premiere date information; title and genre are inconsistent. 
& The genre is a modern drama, which does not match the source entity's Qing Dynasty court setting. \\
\hline
LLM-Result 
& reject 
& \textbf{align ($\times$)} 
& reject 
& reject 
& reject \\
\hline
\end{tabular}
\caption{Case study illustrating how MAD refines alignment decisions through multi-agent debate.
The source entity is $\langle 6270,\ \text{Zijin Jing Lei}\rangle$, and the correct alignment is \textbf{16770}.}
\label{tab:agent-candidate-matrix}
\end{table*}

\begin{table}[t]
\centering
\setlength{\tabcolsep}{2pt} 
\begin{tabular}{l|ccc}
\hline
\multirow{2}{*}{\textbf{Settings}} & \multicolumn{3}{c}{\textbf{DBP15K$_{\text{FR-EN}}$}} \\
& Hits@1 & Hits@10 & MRR \\
\hline
AgentEA & {0.996} & {1.000} & {0.998} \\
AgentEA(Single-LLM) & {0.981} & {0.993} & {0.982} \\
\hline
\end{tabular}
\caption{Ablation experiments during the debate phase on DBP15K$_{\text{FR-EN}}$ dataset.}
\label{tab:ablation3}
\end{table}

\subsection{Why does MAD outperform single-LLM reasoning in EA?}
\label{case}

MAD outperforms single-LLM reasoning by enabling explicit verification, contradiction, and refinement of alignment decisions, which are difficult to achieve under a single-pass reasoning paradigm.

First, as reported in Tables~\ref{tab:dbp15k-results},~\ref{tab:dual-results}, and~\ref{tab:srprs-results}, AgentEA consistently achieves superior performance across multiple EA benchmark datasets. These results demonstrate that MAD provides robust improvements under diverse graph structures and alignment difficulties.

Second, we conduct an ablation study by replacing the proposed two-stage multi-role debate framework with single-LLM reasoning. As shown in Table~\ref{tab:ablation3}, this substitution leads to a substantial performance degradation, indicating that the performance gains cannot be attributed solely to stronger prompts or additional context, but rather to the debate-based reasoning mechanism itself.

Finally, we present a case study to qualitatively illustrate how MAD improves alignment decisions.
As shown in Table~\ref{tab:agent-candidate-matrix}, although the correct entity (16770) is not ranked highest by embedding similarity, MAD progressively accumulates consistent positive signals across multiple agents, including attribute consistency and neighborhood agreement, while suppressing misleading cues through adversarial critique. In contrast, single-LLM reasoning commits to an early judgment based on partial evidence, leading to an overconfident but incorrect alignment. Through iterative debate and explicit cross-agent verification, MAD is able to challenge incorrect assumptions, explore alternative hypotheses, and converge to a more reliable alignment.

\end{document}